\documentclass[11pt]{llncs}
\pdfoutput=1
\title{Probabilistic Model Checking for \\ Complex Cognitive Tasks}
\subtitle{-- A case study in human-robot interaction --}
\author{Sebastian Junges\inst{1}, Nils Jansen\inst{2}, Joost-Pieter Katoen\inst{1}, Ufuk Topcu\inst{2}}

\authorrunning{Junges, Jansen, Katoen, Topcu}
\institute{RWTH Aachen University, Germany \and The University of Texas at Austin, USA}

\usepackage{amsmath,amssymb,amsfonts}
\usepackage{mathtools}
\usepackage[usenames,dvipsnames]{color}
\usepackage{xcolor}
\usepackage{xspace}
\usepackage{microtype}
\usepackage{todonotes}
\usepackage{colonequals}
\usepackage{wasysym}
 \usepackage{booktabs}

\usepackage{multirow}
\usepackage{multicol}

\usepackage{paralist}

\usepackage{subfigure}
\usepackage{url}
\usepackage{appendix}
\usepackage{graphicx}
\usepackage{float}
\usepackage{algorithm}
\usepackage{listings}
\usepackage{nicefrac}
\usepackage{tikz} 
\usepackage{pgfplots}
\usepackage{wrapfig}

\usepackage{siunitx}

\usetikzlibrary{arrows,decorations.pathmorphing,positioning,fit,trees,shapes,shadows,automata,calc}

\renewcommand{\paragraph}[1]{\par\smallskip\noindent\emph{#1}}

\newcommand{\RR}{\ensuremath{\mathbb{R}}}
\newcommand{\NN}{\ensuremath{\mathbb{N}}}

\newcommand{\humanloc}{\ensuremath{\ell_h}}
\newcommand{\humanangle}{\ensuremath{\alpha_h}}
\newcommand{\humanpos}{\ensuremath{\textsf{pos}_h}}
\newcommand{\humaninit}{\ensuremath{\textsf{init}_h}}
\newcommand{\orientations}{\ensuremath{\textsf{Orient}}}

\newcommand{\ath}{\ensuremath{\gamma_h}}
\newcommand{\dth}{\ensuremath{d_h}}

\newcommand{\Tp}{\ensuremath{\mathit{Tp}}}
\newcommand{\OBST}{\ensuremath{\text{Obst}}}
\newcommand{\LM}{\ensuremath{\text{Litt}}}
\newcommand{\WPT}{\ensuremath{\text{Wpt}}}

\newcommand{\Env}{\ensuremath{\textsf{Env}}}

\newcommand{\loc}{\ensuremath{\text{Loc}}}

\newcommand{\tp}{\ensuremath{\textsf{tp}}}

\newcommand{\postact}[1]{\ensuremath{\textsf{post}_#1}}
\newcommand{\Feat}{\ensuremath{\textsf{Feat}}}
\newcommand{\RelFeat}{\ensuremath{\textsf{RelFeat}}}
\newcommand{\ActiveFeat}{\textsf{PFeat}}

\newcommand{\fu}[1]{\ensuremath{\textsf{fu}_#1}}
\newcommand{\eff}[1]{\ensuremath{\textsf{eff}_#1}}

\newcommand{\XDIM}{\ensuremath{\text{Grid}_x}}
\newcommand{\YDIM}{\ensuremath{\text{Grid}_y}}
\newcommand{\dir}{\ensuremath{\textsf{Dir}}}
\newcommand{\humanActs}{\ensuremath{M_\text{h}}}
\newcommand{\HALEFT}{\ensuremath{\textsf{LEFT}}}
\newcommand{\HASTRAIGHT}{\ensuremath{\textsf{STRAIGHT}}}
\newcommand{\HARIGHT}{\ensuremath{\textsf{RIGHT}}}

\newcommand{\wobjectivemovementValues}[1]{w\objectivemovementValues{#1}}
\newcommand{\objectivemovementValues}[1]{\ensuremath{V^#1(s)}}
\newcommand{\movementValues}{\ensuremath{V(s)}}

\newcommand{\close}{\ensuremath{\textsf{Close}}}
\newcommand{\Objectives}{\text{Objectives}}

\newcommand{\softmax}{\ensuremath{\textsf{softmax}}}
\newcommand{\temp}{\ensuremath{\tau}}

\newcommand{\FOLLOW}{\textsf{FOLLOW}}
\newcommand{\AVOID}{\textsf{AVOID}}
\newcommand{\COLLECT}{\textsf{COLLECT}}

\newcommand{\reward}[1]{\textsf{rew}_#1}
\newcommand{\Distr}{\mathit{Distr}}

\newcommand{\bigland}{\ensuremath{\bigwedge}}

\newcommand{\R}{\mathbb{R}}

\newcommand{\finally}{\lozenge}
\newcommand{\sinit}{s_{\mathit{I}}} % initial state of DTMC/MDP
\newcommand{\mdp}{\mathcal{M}}
\newcommand{\probmdp}{\mathcal{P}}
\newcommand{\MdpStates}[1][]{\ensuremath{S{#1}}}
\newcommand{\MdpInit}[1][]{\ensuremath{\mdp{#1}=(\MdpStates[#1],\sinit{#1},\Act,\probmdp})}

\newcommand{\act}[1][a]{\alpha} % single action of MDP
\newcommand{\Act}{\mathit{Act}}        % action set of MDP

\newcommand{\p}{\ensuremath{\mathbb{P}}}

\newcommand{\reachProp}[2]{\ensuremath{\p_{\leq #1}(\finally #2)}}
\newcommand{\reachProplT}{\ensuremath{\reachProp{\lambda}{T}}}

\newcommand{\rew}{\ensuremath{\mathrm{rew}}}
\newcommand{\e}{\ensuremath{\mathbb{E}}}

\newcommand{\expRewProp}[2]{\ensuremath{\e_{\leq #1}(\eventually #2)}}

\newcommand{\eventually}{\ensuremath{\lozenge}}

\newcommand{\spOne}{\ensuremath{S_{\circ}}}
\newcommand{\spTwo}{\ensuremath{S_{\Box}}}
\newcommand{\pOne}{\ensuremath{{\circ}}}
\newcommand{\pTwo}{\ensuremath{{\Box}}}
\newcommand{\probsg}{\probmdp}
\newcommand{\psgGeneric}{\ensuremath{\mathfrak{M}}}

\newcommand{\sgInitGeneric}[1][]{\ensuremath{\psgGeneric{#1}=(S{#1},\sinit{#1},\Act,\probsg{#1})}}

\newcommand{\ltp}[1][L]{\mathcal{L}}   % linear-time property 
\newcommand{\eg}{e.\,g.\xspace}

\newcommand{\bddstate}[1]{s_{\mathbb{B}}{}}

\newcommand{\tool}[1]{\texttt{#1}\xspace}

\newcommand{\prism}{\tool{PRISM}}
\DeclareRobustCommand{\cpp}
{\valign{\vfil\hbox{##}\vfil\cr
   \textsf{C\kern-.1em}\cr
   $\hbox{\fontsize{\sf@size}{0}\textbf{+\kern-0.05em+}}$\cr}%
}

\selectcolormodel{cmy}

\definecolor{lgrey}{rgb}{0.8,0.8,0.8}
\definecolor{grey}{rgb}{0.5,0.5,0.5}

\definecolor{lightblue}{rgb}{0.8,0.8,1.0}
\definecolor{lightred}{rgb}{1.0,0.8,0.8}
\definecolor{lightgreen}{rgb}{0.8,1.0,0.8}
\definecolor{angrygreen}{cmyk}{0.279,0,0.91,0.08}
\definecolor{lightred}{rgb}{1.0,0.8,0.8}
\definecolor{pink}{rgb}{1.0,0.1,1.0}

\definecolor{prismgreen}{HTML}{009900}
\definecolor{prismred}{HTML}{cc0000}
\definecolor{prismblue}{HTML}{0000cc}

\lstdefinelanguage{Prism}{
        basicstyle=\color{prismred}\scriptsize\ttfamily,
        literate=*	{0}{{\textcolor{prismblue}{0}}}{1}
			{1}{{\textcolor{prismblue}{1}}}{1}
			{2}{{\textcolor{prismblue}{2}}}{1}
			{3}{{\textcolor{prismblue}{3}}}{1}
			{4}{{\textcolor{prismblue}{4}}}{1}
			{5}{{\textcolor{prismblue}{5}}}{1}
			{6}{{\textcolor{prismblue}{6}}}{1}
			{7}{{\textcolor{prismblue}{7}}}{1}
			{8}{{\textcolor{prismblue}{8}}}{1}
			{9}{{\textcolor{prismblue}{9}}}{1}
			{.0}{{\textcolor{prismblue}{.0}}}{2}
			{.1}{{\textcolor{prismblue}{.1}}}{2}
			{.2}{{\textcolor{prismblue}{.2}}}{2}
			{.3}{{\textcolor{prismblue}{.3}}}{2}
			{.4}{{\textcolor{prismblue}{.4}}}{2}
			{.5}{{\textcolor{prismblue}{.5}}}{2}
			{.6}{{\textcolor{prismblue}{.6}}}{2}
			{.7}{{\textcolor{prismblue}{.7}}}{2}
			{.8}{{\textcolor{prismblue}{.8}}}{2}
			{.9}{{\textcolor{prismblue}{.9}}}{2}
			{[}{{\textcolor{black}{[}}}{1}
			{]}{{\textcolor{black}{]}}}{1}
			{+}{{\textcolor{black}{+}}}{1}
			{-}{{\textcolor{black}{-}}}{1}
			{=}{{\textcolor{black}{=}}}{1}
			{<}{{\textcolor{black}{<}}}{1}
			{>}{{\textcolor{black}{>}}}{1}
			{\&}{{\textcolor{black}{\&}}}{1}
			{|}{{\textcolor{black}{|}}}{1}
			{:}{{\textcolor{black}{:}}}{1}
			{;}{{\textcolor{black}{;}}}{1}
			{(}{{\textcolor{black}{(}}}{1}
			{)}{{\textcolor{black}{)}}}{1}
			{..}{{\textcolor{black}{..}}}{2},
        keywords= {bool,C,ceil,const,ctmc,double,dtmc,endinit,endmodule,endrewards, endsystem,F,false,floor,formula,G,global,I,init,int,label,max,mdp,min,module,nondeterministic,P,Pmin,Pmax,prob,probabilistic,R,rate,rewards,Rmin,Rmax,S,stochastic,system,true,U,X},
        keywordstyle={\bfseries\color{black}},
        numberstyle=\footnotesize\color{black},
        comment=[l] {//}, morecomment=[s]{/*}{*/},
        commentstyle= \color{prismgreen},
        tabsize=4,
        captionpos=b,
        escapechar=@
}

\makeatletter
\pgfdeclareshape{document}{
\inheritsavedanchors[from=rectangle] % this is nearly a rectangle
\inheritanchorborder[from=rectangle]
\inheritanchor[from=rectangle]{center}
\inheritanchor[from=rectangle]{north}
\inheritanchor[from=rectangle]{south}
\inheritanchor[from=rectangle]{west}
\inheritanchor[from=rectangle]{east}
\backgroundpath{% this is new
\southwest \pgf@xa=\pgf@x \pgf@ya=\pgf@y
\northeast \pgf@xb=\pgf@x \pgf@yb=\pgf@y
\pgf@xc=\pgf@xb \advance\pgf@xc by-10pt % this should be a parameter
\pgf@yc=\pgf@yb \advance\pgf@yc by-10pt
\pgfpathmoveto{\pgfpoint{\pgf@xa}{\pgf@ya}}
\pgfpathlineto{\pgfpoint{\pgf@xa}{\pgf@yb}}
\pgfpathlineto{\pgfpoint{\pgf@xc}{\pgf@yb}}
\pgfpathlineto{\pgfpoint{\pgf@xb}{\pgf@yc}}
\pgfpathlineto{\pgfpoint{\pgf@xb}{\pgf@ya}}
\pgfpathclose
\pgfpathmoveto{\pgfpoint{\pgf@xc}{\pgf@yb}}
\pgfpathlineto{\pgfpoint{\pgf@xc}{\pgf@yc}}
\pgfpathlineto{\pgfpoint{\pgf@xb}{\pgf@yc}}
\pgfpathlineto{\pgfpoint{\pgf@xc}{\pgf@yc}}
}
}
\makeatother

\tikzset{
    human/.style 2 args={
        draw=black,
        shape=isosceles triangle,
        fill=#2,
        minimum height=0.65cm,
        minimum width=0.5cm,
        shape border rotate=0,
        rotate=#1,
        isosceles triangle stretches,
        inner sep=0pt
    },
    obstacle/.style={
    	draw=red,
    	cross out,
    	very thick,
    	minimum size=0.5cm, 
    	inner sep=0pt, outer sep=0pt
    },
    landmark/.style={
    	rectangle,
    	draw=blue,
    	fill=blue!70,
    	minimum size=0.5cm,
    	inner sep=0pt, outer sep=0pt
    },
    waypoint/.style={
    	circle,
    	draw=black,
    	fill=black!70,
    	minimum size=0.3cm,
    	inner sep=0pt, outer sep=0pt
    }
    }

\allowdisplaybreaks
\begin{document}
\maketitle

\begin{abstract}
This paper proposes to use probabilistic model checking to synthesize
optimal robot policies in multi-tasking autonomous systems that are
subject to human-robot interaction. Given the convincing empirical
evidence that human behavior can be related to reinforcement models, we
take as input a well-studied Q-table model of the human behavior for 
flexible scenarios. We first describe an automated procedure to distill a
Markov decision process (MDP) for the human in an arbitrary but fixed scenario. The
distinctive issue is that -- in contrast to existing models --
under-specification of the human behavior is included. Probabilistic
model checking is used to predict the human's behavior. Finally, the MDP
model is extended with a robot model. Optimal robot policies are
synthesized by analyzing the resulting two-player stochastic game.
Experimental results with a prototypical implementation using PRISM show
promising results.
\end{abstract}

\section{Introduction}\label{sec:intro}

Verification and design for autonomous systems that work with humans account for the human's capabilities, preferences and limitations by embedding behavioural models of \emph{humans}. 
With increasing capabilities to monitor humans in dynamic, possibly mixed-reality environments, data-driven modeling enables to encode such data into behavioural models. 
Moreover, \emph{reinforcement learning}~\cite{sutton1998reinforcement} (RL) sufficiently describes quantitative aspects of human behaviour when solving complicated tasks in realistic environments~\cite{barto199511,daw2006computational}. 
Basically, RL comprises algorithms addressing the optimal control problem through learning, i.e., an agent---the human---learns how to solve a task based on repeated interaction with an environment. 
So--called \emph{Q-tables} store quantitative information about possible choices of the human. 
We consider this information the \emph{data} describing human behaviour.

\begin{figure}[t]
	\centering
%	\subfigure[Simulation environment showing an avatar and head centered views with target detection.]{
%		\includegraphics{pics/schemeA}\label{fig:visioA}
%	}
%		\subfigure[Division of state space into the modular tasks (i) approach target, (ii) avoid obstacle, and (iii) follow walkway.]{
		\includegraphics{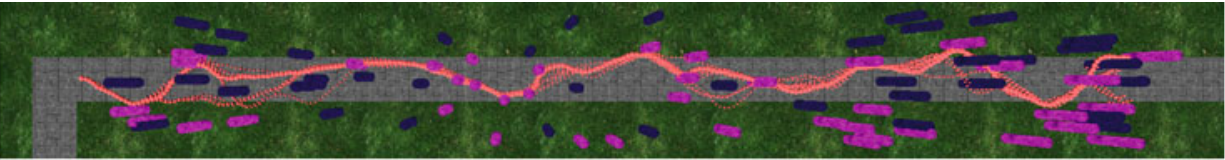}\label{fig:visioB}
%	}
	\caption{Visio-motor setting with different tasks while walking down a sidewalk shown in a simulation environment. The state space is divided into the modular tasks (i) approach targets (purple), (ii) avoid obstacles (blue), and (iii) follow walkway (grey). This picture is taken from~\cite{rothkopf2013modular} with permission from Ballard.}
	\label{fig:visio}
%	\vspace*{-0.4cm}
\end{figure}

Consider the \emph{visio-motor} setting from~\cite{rothkopf2013modular} in Fig.~\ref{fig:visio}.
A human walks down a sidewalk and shall attend to three modular tasks: avoiding obstacles (purple), approaching targets (blue), or following the walkway (grey). 
RL generates Q-tables for the individual tasks; each table quantifies available choices that ultimately lead to completing the task.
To build an accurate and general model of observed human behaviour for different tasks, \emph{inverse reinforcement learning}   (IRL)~\cite{ng2000algorithms} assigns \emph{weights} describing preferences over these tasks. 
It might, \eg, be more important to avoid an obstacle than to approach a target.
In particular, IRL connects data sets about human behaviour---either observed over time or obtained by RL---to a general model describing how a human typically behaves in presence of different tasks.
A large class of human behavioural models is covered by a set of Q--tables together with weights obtained by the methods in~\cite{rothkopf2013modular}. 
The given tables describe behaviour for generic scenarios in the sense that they take into account distances to features such as obstacles or targets instead of their concrete position. 

\begin{figure}[t]
\scalebox{0.9}{
\input{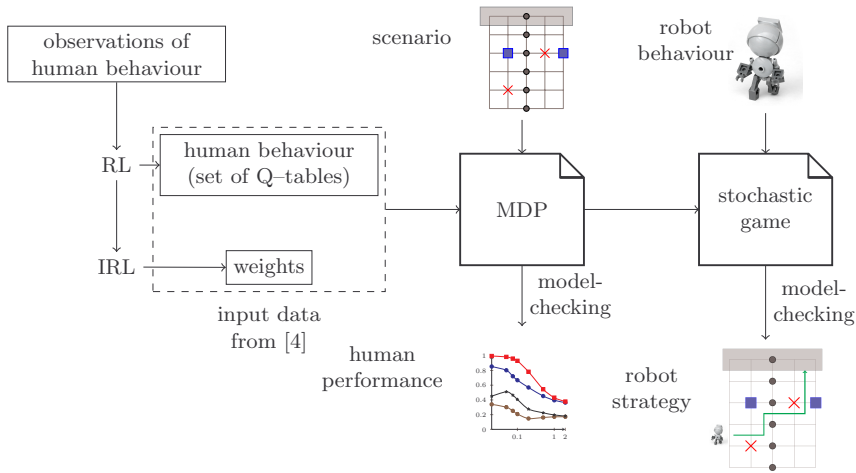}
}
%\vspace{-8mm}
\caption{Overview of our approach to verify human-robot interaction.}
\label{fig:tool}
%	\vspace*{-0.4cm}
\end{figure}

This paper proposes \emph{probabilistic model checking}~\cite{Kat16} to analyse human behaviour models described by weighted Q--tables\footnote{
 A high-level conceptual view in the form of an extended abstract is given in \cite{cdcas}. All technical content, implementation details, experiments and lessons learned are novel.}
For an arbitrary \emph{concrete scenario}, a Markov decision process (MDP)~\cite{Put94} is generated automatically, see Fig.~\ref{fig:tool}.
Intuitively, this MDP describes all possible human behaviours for this concrete scenario based on the behavioural model.
The distinctive issue is that---in contrast to existing models---under-specification of the human behaviour is included.
We assess the performance of the human for the scenario as well as properties of the human model itself by employing MDP model checking
as supported by \tool{PRISM}~\cite{KNP11}, \tool{StORM}, and \tool{iscasMc}~\cite{iscasmc}. 
We then assess robot behaviour in the presence of possibly disturbing humans.
This is done by combining the human model with a robot-MDP model.
The joint human-robot interaction model is a stochastic two-player game (SG)~\cite{DBLP:journals/iandc/Condon92}.
We synthesise optimal policies for the robot under the human behaviour using SG model-checking with \tool{PRISM-Games}~\cite{DBLP:conf/tacas/KwiatkowskaPW16}.

We stress that our approach is applicable to any human behaviour models described by weighted Q--tables.
The approach is evaluated on an existing model by Rothkopf \emph{et al.} for visio-motor tasks~\cite{rothkopf2010credit,rothkopf2013modular,matt_visiomotor}. 
The main bottleneck is to handle the textual MDP description of the human behaviour of over 80,000 lines of \tool{PRISM}-code.
We thoroughly analysed MDPs of $10^6$ states induced by a $20{\times}20$ grid scenario. 
The human-interaction model for a $8{\times}8$ grid scenario constitute a SG with $1.6{\cdot}10^7$~states, its generation---in absence of a symbolic engine~\cite{par02,baier_symbolic_modelchecking} for SGs---takes over twenty hours; analysing maxmin reachability probabilities takes three hours. The SGs have a noticeably more complex structure than benchmarks in~\cite{CFK+13} and offer new challenges to probabilistic verification.

%
%
%\paragraph{Probabilistic modeling and verification.}
% We demonstrate how a class of human behavioral models can be translated into Markov decision processes (MDPs). We then utilize probabilistic model checkers to, e.g., assess the expected performance of the human and the effects of the modeling choices made prior to the verification step. Furthermore, we combine the human behavioral model with a model for a robot into a stochastic two-player game and synthesize control protocols for the robot. While the approach is general and flexible, we evaluate it on an existing model for visiomotor tasks. 
% 
% \paragraph{Contribution -- Lessons learned.}
%\nj{state clearly the lessions learned, why it is important to investigate this case study.}
%

\section{Preliminaries}
\label{sec:preliminaries}
%For a set $X$, let $2^{X}$ denote the power set of $X$. 
%A \emph{probability distribution} over a finite or countably infinite set $\distDom$ is a function $\distFunc\colon\distDom\rightarrow\Ireal$ with $\sum_{\distDomElem\in\distDom}\distFunc(\distDomElem)=\distFunc(\distDom)=1$. 
%In this paper, all probabilities are taken from $\Q$.
%Let the set of all distributions on $\distDom$ be denoted by $\Distr(\distDom)$.
%The set $\supp(\distFunc)=\{x\in\distDom \mid \distFunc(x)>0\}$ is the \emph{support} of $\distFunc \in \Distr(\distDom)$. 
%If $\mu(x)=1$ for $x\in\distDom$ and $\mu(y)=0$ for all $y\in\distDom\setminus\{x\}$, $\mu$ is called a \emph{Dirac distribution}.
%\nj{remove what is not needed}
%\subsection{Probabilistic Models}\label{sec:preliminaries:models}
%We introduce three types of \emph{probabilistic models}.
%, which can be seen as transition systems (with a possible partition of the state space into two sets) where the transitions are labeled with probabilities. 
%
In this section, we give a short introduction to models, specifications, and our notations; for details we refer to~\cite[Ch.\ 10]{BK08}.
\begin{definition}[Probabilistic models]\label{def:prob_model}
A \emph{stochastic game (SG)} is a tuple $\sgInitGeneric$ with a finite set $S$ of states such that $S = \spOne\uplus\spTwo$, an initial state $\sinit \in S$, a finite set $\Act$ of actions, and a transition function $\probmdp \colon S \times \Act \times S \rightarrow [0,1]$ and $\sum_{s'\in S}\probmdp(s,\act,s') \in \{0, 1\}\quad \forall s\in S, a \in \Act$.

\begin{compactitem}
\item
$\psgGeneric$ is a \emph{Markov decision process (MDP)} if $\spOne=\emptyset$ or $\spTwo=\emptyset$.
\item
MDP $\psgGeneric$ is a \emph{Markov chain (MC)} if $|\Act(s)|=1$ for all $s \in S$.
\end{compactitem}
\end{definition}
%%
%We refer to MCs by $\dtmc$, to MDPs by $\mdp$ and to SGs by $\psg$.
We refer to MDPs by $\mdp$.
SGs are two-player stochastic games with players $\pOne$ and $\pTwo$ having states in $\spOne$ and $\spTwo$, respectively.
Players \emph{nondeterministically} choose an action at each state; successors are determined \emph{probabilistically} according to transition probabilities.
%For all states $s\in S$, we assume the set of \emph{enabled} actions $\Act(s) = \{\act \in \Act \mid \exists s'\in S.\,\probmdp(s,\act,s') \neq 0\}$ to be non-empty, \ie, there are no deadlock states.
%
MDPs and MCs are one- and zero-player stochastic games, respectively.
As MCs have one action at each state, we omit this action and write $\probmdp(s,s')$. 
For analysis, w.l.o.g.\ we assume that in each state there is at least one action available.

Probabilistic models are extended with \emph{rewards} (or costs) by adding a \emph{reward function} $\rew\colon S \rightarrow \R_+$ which assigns rewards to states of the model.
Intuitively, the reward $\rew(s)$ is earned upon leaving the state $s$.
%For pMCs, we can omit the actions and consider reward functions of the form $\rewFunction \colon S \times S \rightarrow \Re_{\geq 0}$ instead.
%In fact, rewards could also be considered parametric in which case they associate a polynomial to a transition\sj{Are we using transition or state rewards?}.\footnote{This is used in one of our benchmarks later on.}
%
%\medskip\noindent\emph{Schedulers.}
Nondeterministic choices of actions in SGs and MDPs are resolved \emph{schedulers}; 
%\footnote{Also referred to as adversaries, strategies, or policies.}.
here it suffices to consider memoryless deterministic schedulers~\cite{Var85}.
%, of the form $\sched\colon S\rightarrow\Act$ with $\sched(s)\in \Act(s)$ for all $s\in S$. 
%
%\begin{definition}{\bf (Scheduler)}\label{def:scheduler}
%	A \emph{scheduler} for MDP $\mdpInit$ is a function $\sched\colon S\rightarrow\Act$ with $\sched(s)\in \Act(s)$ for all $s\in S$.  
%\end{definition}
%
%Let $\Sched{\mdp}$ denote the set of all schedulers for $\mdp$.
Resolving all nondeterminism in SGs or MDPs yields \emph{induced Markov chains}; note that for SG we need individual schedulers for each player. 
%\begin{definition}{\bf (Induced MC)}\label{def:induced_dtmc} 
%	For MDP $\mdpInit$ and scheduler $\sched\in\Sched{\mdp}$, the \emph{MC induced by $\mdp$ and $\sched$} is $\mdp^\sched=(S, \sinit,\probmdp^\sched)$ where
%	\begin{align*}
%		\probmdp^\sched(s,s')= \probmdp(s,\sched(s),s') \quad \mbox{ for all } s,s'\in S\ .
%	\end{align*} 
%	%% For MDPs equipped with rewards, we define: $\rewFunction(s,s') = \rewFunction(s,\sched(s),s')$.
%\end{definition}
%%
%Intuitively, the transition probabilities in $\mdp^\sched$ are obtained \wrt the choice of actions by the scheduler. 
%We apply the same notions to define the satisfaction for expected reward properties.
%
%For $\spOne$ and $\spTwo$ we need schedulers $\sched \in \Sched[\pOne]{\sg}$ and $\altsched \in \Sched[\pTwo]{\sg}$ of the form $\sched \colon {\spOne}\rightarrow\Act$ and $\altsched\colon \spTwo\rightarrow\Act$. The induced MC $\sg^{\sched, \altsched}$ of a SG $\sg$ with schedulers $\sched$ and $\altsched$ for both players is defined analogously to the one for MDPs.

%%

%\subsection{Properties of Interest}

As specifications we consider \emph{reachability} and \emph{expected reward} properties. 
%For MC $\dtmc$ with state space $S$, let $\reachPrs{\dtmc}{s}{T}$ denote the probability to reach a set of target states $T \subseteq S$ from  state $s\in S$ within $\dtmc$
%; simply $\reachPrT[\dtmc]$ refers to this specific probability for the initial state $\sinit$.
%We use a standard probability measure on infinite paths through an MC as defined in~\cite[Ch.\ 10]{BK08}.
A {reachability property} asserts that a set $T \subseteq S$ of target states is to be reached from the initial state with probability at most $\lambda\in [0,1]$, denoted $\reachProplT$. 
%The property is satisfied by $\dtmc$, written $\dtmc \models \reachPropSymbol$, iff $\reachPrT[\dtmc]\leq\lambda$.
%(Comparisons like $<$, $>$, and $\geq$ are treated in a similar way.)
%%The reward of a path through an MC $\dtmc$ until $T$ is the sum of the rewards of the states visited along on the path before reaching $T$.
%%The expected reward of a finite path is given by its probability times its reward.
%Given $\reachPrT[\dtmc] = 1$, the expected reward of reaching $T \subseteq S$, is the sum of the expected rewards of all paths to reach $T$.
%In case $\reachPrT[\dtmc] < 1$, we set $\expRewT[\dtmc] = \infty$.\sj{As we do not go into technical details, this sentence seems superfluous}
Analogously, an expected reward property bounds the expected reward of reaching $T$ by a threshold $\kappa \in \R$, denoted $\expRewProp{\kappa}{T}$.%% \tq{Mention the case where $T$ is not reachable with probability 1? This case becomes critical (and ugly) when we want a formal proof for rewards later..}
%% We will also consider properties of the form $\neg \varphi$ with the usual meaning: $\dtmc \models \neg \varphi$ iff $\dtmc \not\models \varphi$.\nj{I would say this is clear.}
%We often omit the superscript $\pdtmc$ if it is clear from the context. 
%%We refer to~\cite{BK08} for more details on probability measures and reachability properties 
%% \tq{$\ge,  >,  <$ are also possible}
%
%\medskip\noindent\emph{Properties on nondeterministic models.}

For MDPs and SGs, properties have to hold for all possible schedulers on the corresponding induced MCs. Verification can be performed by computing maximal (or minimal) probabilities or expected rewards to reach target states using standard techniques, such as linear programming, value iteration, or policy iteration~\cite{Put94}.

\section{Description of the cognitive model}
This section describes the scenario of the cognitive model used as case study.
It also define a formalisation that can be applied to similar case studies. 
This provides the basis to obtain the underlying representation as an MDP and---incorporating robot movement---an SG.

\begin{figure}[t]
\centering
\subfigure[From \cite{rothkopf2013modular} \scriptsize{(with permission)}]{
\phantom{a}
\includegraphics[scale=0.69]{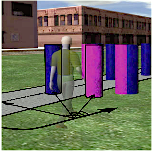}
\phantom{a}
}
\subfigure[Toy-scenario]{
\scalebox{0.7}{
	\begin{tikzpicture}[label distance=-1pt]
\node[waypoint, label={$\WPT$}] at (0.2, 4.6) {};
\node[landmark, label={$\LM$}] at (1.1, 4.6) {};
\node[obstacle, label={$\OBST$}] at (2.1, 4.6) {};
\node[human={180}{green!80}, opacity=0.4, label={270:$\humaninit$}] at (3.1,4.6) {};
\node[rectangle, fill=black!40!white, inner sep=4pt, opacity=0.5, label={Goal}] at (3.9, 4.6) {};
%\draw[fill=green!40!white, opacity=0.6] (3.9, -0.1) rectangle (4.1, 0.1);
\draw[step=1cm,gray,very thin] (0,0) grid (4,4);
%\draw[fill=black!40!white, opacity=0.5] (-0.3, 3.5) rectangle (4.3, 4.5);
\node[obstacle, label={260:$f_5$}] at (3,3) {};
\node[obstacle, label={260:$f_6$}] at (0,2) {};
\node[obstacle, label={260:$f_7$}] at (0,4) {};
\node[landmark, label={260:$f_8$}] at (1,3) {};
\node[landmark, label={260:$f_9$}] at (4,3) {};
%%
%\node[waypoint] at (4,0) {};
\node[waypoint, label={260:$f_1$}] at (2,1) {};
\node[waypoint, label={260:$f_2$}] at (2,2) {};
\node[waypoint, label={260:$f_3$}] at (2,3) {};
\node[waypoint, label={260:$f_4$}] at (2,4) {};
\node[human={90}{green!80}, opacity=0.4] at (2,0) {};
\draw[fill=black!40!white, inner sep=4pt, opacity=0.5] (-0.3, 3.7) rectangle (4.3, 4.3);
\end{tikzpicture}}
\label{Fig:ToyScenario}
}
\subfigure[Example situation]{
\scalebox{0.7}{
\begin{tikzpicture}
\draw[step=1cm,gray,very thin] (0,0) grid (4,4);
\node[obstacle] at (3,3) (f5) {};
\node[obstacle] at (0,2) (f6) {};
\node[obstacle, opacity=0.7] at (0,4) (f7) {};
\node[landmark] at (1,3) (f8) {};
\node[landmark, opacity=0.7] at (4,3) (f9) {};

\node[waypoint] at (2,2) (f2) {};
\node[waypoint, opacity=0.7] at (2,3) (f3) {};
\node[waypoint, opacity=0.7] at (2,4) (f4) {};
\node[human={90}{green!80}] at (2,1) (h) {};

\draw[fill=black!40!white, inner sep=4pt, opacity=0.5] (-0.3, 3.7) rectangle (4.3, 4.3);

\draw[->, thick] (h) -- (f2);

\draw[->, thick] (h) -- (f5);
\draw[->, thick] (h) -- (f6);
\draw[->, thick] (h) -- (f8);

\end{tikzpicture}
}	\label{Fig:ToyScenSituation}
}
\caption{Graphical representation of the grid worlds.}
\end{figure}
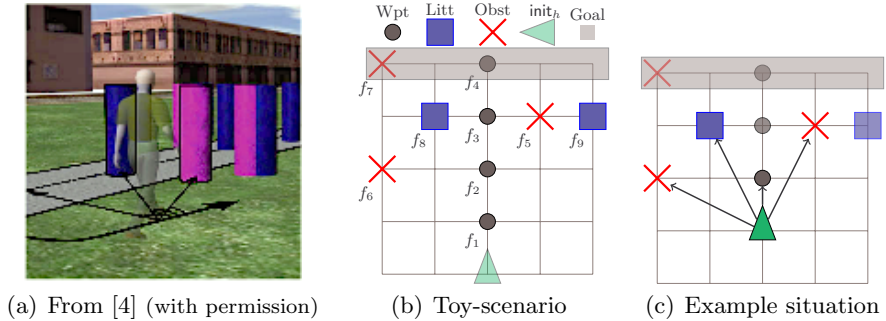

\paragraph{General scenario.}
We consider a scenario involving a human agent going over a sidewalk encountering \emph{objects} like \emph{obstacles} and \emph{litter}, see Fig.~\ref{fig:visioB}.
The human is given three modular objectives: while \textsf{FOLLOW} a sidewalk (represented as line) to get to the other side, she should \textsf{AVOID} walking into obstacles and aim to \textsf{COLLECT} litter.
%
%\begin{description}
%\item{\textsf{FOLLOW}:} The human should follow a line; the line is an abstraction of the sidewalk the human is supposed to walk over to the other side. \nj{unclear?}
%\item{\textsf{AVOID}:} The human should not walk into obstacles. 
%\item{\textsf{COLLECT}:} The human should aim to collect litter. 
%\end{description}
The line to follow is abstracted as a set of \emph{waypoints}. 
Waypoints and objects such as obstacles and litter are called \emph{features} of the scenario.
As obstacles can be run over, litter can be collected and waypoints can be visited, these features are either being \emph{present} or \emph{disappeared}.
For analysis purposes, we mark specific regions of the environment as \emph{goal-areas}.
We assume that all features are initially present and that disappeared features remain absent. 
A toy-scenario is given in Fig.~\ref{Fig:ToyScenario}. The general scenario is abstracted to a discrete state model to allow to characterise human behaviour in the model. It consists of two parts: a (mostly) static environment and the dynamic human.
The behaviour is determined by the movement-values assigned to each movement. These values are then translated into a probability distribution over the movements\footnote{We use the term movement to avoid confusion with actions in MDPs.}.

\subsection{Formal model} 
Let us describe how the human interacts with its environment.
The environment consists of a two-dimensional grid and its features, where features have a type in $\Tp = \{\OBST,\LM,\WPT\}$.
\begin{definition}[Environment]
An \emph{environment} $\Env=(\loc, \Feat)$ consists of a finite set of \emph{locations} $\loc$ with \[\loc = \{(x, y) \mid x \in [0, \XDIM]~y \in [0, \YDIM]\} \qquad\text{ for }\XDIM, \YDIM \in \NN, \] 
and a set of \emph{features} $\Feat \subseteq \Tp \times \loc$.
A \emph{feature} $f = (\tp_f, \ell_f) \in \Feat$ consists of a \emph{type} and a \emph{(feature-)location}.
\end{definition} 
Features are partitioned according to their type: $\Feat = \Feat_\OBST  \cup  \Feat_\LM \cup \Feat_\WPT$, such that $\Feat_\tp \subseteq \{ \tp \} \times \loc$.
\begin{example}
Consider our running example in Fig.~\ref{Fig:ToyScenario}. The environment is given as $\Env = (\loc, \Feat)$ 
with $ \loc = \{ (x,y) \mid x \in [0,4]~y \in [0,4] \}$, and
\begin{align*}
 \Feat & = \{ f_i = (\WPT, (2,i)) \mid i \in \{ 1 \hdots 4\}~\} \\ & \cup \{ f_5 = (\OBST, (3, 3)), f_6 = (\OBST, (0,2)), f_7 = (\OBST, (0,4)) \} \\ & \cup \{ f_8 = (\LM, (1,3)), f_9 = (\LM, (4,3)) \}
\end{align*}
\end{example}

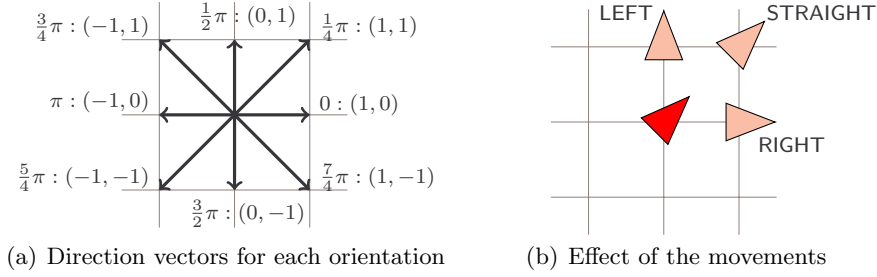
\begin{figure}[t]
\centering
\subfigure[Direction vectors for each orientation]{
\begin{tikzpicture}
\draw[step=1cm,gray,very thin] (-1.5,-1.5) grid (1.5,1.5);	
\draw[->, very thick] (0,0) -- node[pos=1.0, right, yshift=3.8] {\scriptsize{$0: (1,0)$}} (1,0);
\draw[->, very thick] (0,0) -- node[pos=1.0, right, yshift=4.8] {\scriptsize{$\frac{1}{4}\pi: (1,1)$}} (1,1);
\draw[->, very thick] (0,0) -- node[pos=1.0, above, xshift=4.8] {\scriptsize{$\frac{1}{2}\pi: (0,1)$}} (0,1);
\draw[->, very thick] (0,0) -- node[pos=1.0, left, yshift=4.8] {\scriptsize{$\frac{3}{4}\pi: (-1,1)$}} (-1,1);
\draw[->, very thick] (0,0) -- node[pos=1.0, left, yshift=4.8] {\scriptsize{$\pi: (-1,0)$}} (-1,0);
\draw[->, very thick] (0,0) -- node[pos=1.0, left, yshift=4.8] {\scriptsize{$\frac{5}{4}\pi: (-1,-1)$}} (-1,-1);
\draw[->, very thick] (0,0) -- node[pos=1.0, below, xshift=4.8] {\scriptsize{$\frac{3}{2}\pi: (0,{-}1)$}} (0,-1);
\draw[->, very thick] (0,0) -- node[pos=1.0, right, yshift=4.8] {\scriptsize{$\frac{7}{4}\pi: (1,{-}1)$}} (1,-1);

\end{tikzpicture}

\label{Fig:Orientations}
}
\subfigure[Effect of the movements]{
\begin{tikzpicture}
\draw[draw=white, very thin] (-2.6, -1.5) grid (2.6, 1.5);
\draw[step=1cm,gray,very thin] (-1.5,-1.5) grid (1.5,1.5);	
\node[human={45}{red}] at (0.0, 0.0) {};
\node[human={0}{red!30}] at (1.0, 0.0) {};
\node[] at (1.7, -0.3) {\scriptsize{$\HARIGHT$}};
\node[human={45}{red!30}] at (1.0, 1.0) {};
\node[] at (2.1, 1.45) {\scriptsize{$\HASTRAIGHT$}};
\node[human={90}{red!30}] at (0.0, 1.0) {};
\node[] at (-0.5, 1.45) {\scriptsize{$\HALEFT$}};
\end{tikzpicture}
\label{Fig:HumanMovementEffect}
}	
\caption{Orientations and human movement.}
\end{figure}

\paragraph{Human.}
The human $h$ is represented by its \emph{position} $\humanpos = (\humanloc, \humanangle)$ which is a pair of a location $\humanloc$ and an orientation $\humanangle$. 
An orientation has 8 possible directions, i.e. $\humanangle \in \orientations = \{ i \cdot \frac{1}{4}\pi \mid i \in [0, 7]\}$. 
We assume that the human starts in position $\humaninit$.

\paragraph{Human movement.}
For each direction, let the associated direction vector $\dir\colon \orientations \rightarrow \{-1, 0, 1\}^2 \setminus \{(0,0)\}$, see Fig.~\ref{Fig:Orientations}. 
Human movements $\humanActs = \{ \HALEFT, \HASTRAIGHT, \HARIGHT \}$, see Fig.~\ref{Fig:HumanMovementEffect}, have associated changes in angle of $\beta = {-}\frac{1}{4}\pi$, $0$, or $\frac{1}{4}\pi$.  
To be precise, the \emph{post-position} $\postact{m}(\humanpos)$ 
%\begin{definition}[Post-position of a movement]
for human position $\humanpos = (\humanloc, \humanangle)$ and movement $m$ with $\beta_m$, is given by:
\[\postact{m}(\humanpos) = (\humanloc + \dir(\humanangle + \beta_m),\quad(\humanangle + \beta_m)\;\bmod\; 2\pi).\] 
%\end{definition}
A movement $m$ is \emph{valid in $\humanpos$} if $\postact{m}(\humanpos)$ is a well-defined position, that is, the location of the post-position is on the grid.

\paragraph{Situations.}
The status of a scenario is called a \emph{situation}:
\begin{definition}[Situation]
Let  $\Env=(\loc, \Feat)$ be an environment. A \emph{situation} $s$ is a pair of a human location and the \emph{present features} $(\humanpos, \ActiveFeat) \in \loc \times 2^{\Feat}$. 
\end{definition}
As we see in Sect.~\ref{Sec:FormalMDP}, the situations define the state space of our model.
If the post-position of a movement coincides with a feature, then the feature is marked disappeared. 
Formally, for a valid movement $m$ in  position $\humanpos$, the present features $\ActiveFeat$ are updated as follows: \[\fu{m}(s) = \ActiveFeat \setminus \{ (\tp, \postact{m}(\humanpos)) \} \subseteq \Feat.\]
 The effect of a movement of the human is as follows.
\begin{definition}[Human movement effect]
	The effect of a human movement $m$ in a situation $s=(\humanpos, \ActiveFeat)$ is the situation \[\eff{m}(s) = (\postact{m}(\humanpos), \fu{m}(s)).\]
\end{definition}

\paragraph{Distance and angle.}
Let $f$ be a feature and $\humanpos = (\humanloc, \humanangle)$ the position of the human $h$.
The distance between $h$ and $f$ is given by the Euclidean norm, i.e.\ $\dth(f) = \| \humanloc - \ell \|_2$, and $\ath(f)$ denotes the signed angle between the human orientation and $\humanloc - \ell$. 

\begin{example}
In the situation $s=(\humanpos, \ActiveFeat)$ depicted in Fig.~\ref{Fig:ToyScenSituation}, $\eff{\HALEFT}(s) = (((1,2), \frac{3}{4}\pi), \ActiveFeat \setminus \{ f_2 \})$.
The distance of the human $h$ to $f_9$ at $(4,3)$ is $\dth(f_9) = \|(4,3) - (2,1)\|_2 = 2\sqrt{2}$, the angle is $\ath(f_9) = -\frac{1}{4}\pi$.
\end{example}
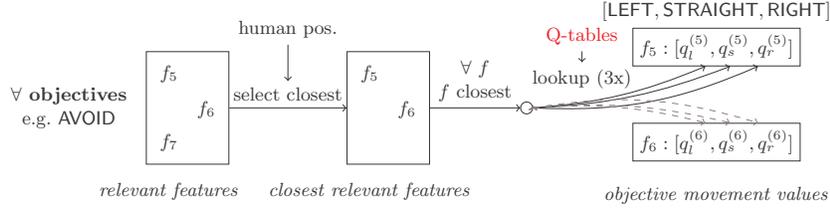
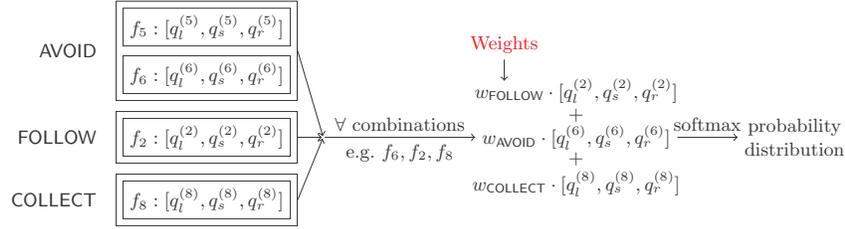
\begin{figure}[t]\subfigure[The computation of movement values]{
\scalebox{0.8}{
\begin{tikzpicture}
	\node[] (f1) {$f_5$};
	\node[below=0.2cm of f1] (f2) {};
	\node[below right=0.12cm of f1] (f4) {$f_6$};
	\node[below=0.2cm of f2] (f3) {$f_7$};
	\node[below=0.3cm of f3, align=center] (ftext) {\emph{relevant features}};
	\node[fit=(f1)(f2)(f3)(f4), draw] (fbox) {};
	
	\node[right=2.8cm of f1] (g1) {$f_5$};
	\node[below=0.2cm of g1] (g2) {};
	\node[below right=0.12cm of g1] (g4) {$f_6$};
	\node[below=0.2cm of g2] (g3) {\phantom{$f_3$}};
	\node[below=0.3cm of g3] (gtext) {\emph{closest relevant features}};
	\node[fit=(g1)(g2)(g3)(g4), draw] (gbox) {};

	\node[right=4.1cm of g1, yshift=0.44cm, draw,label={$[\HALEFT, \HASTRAIGHT, \HARIGHT]$}] (feat1) {$f_5: [q^{(5)}_l, q^{\scriptsize{(5)}}_s, q^{(5)}_r]$};
	
	\node[right=4.1cm of g1, yshift=-1.14cm, draw] (feat2) {$f_6: [q^{(6)}_l, q^{(6)}_s, q^{(6)}_r]$};
	
%	\node[right=3.5cm of g1, yshift=-0.44cm] (h21) {$a^s_1$};
%	\node[below right=0.12cm of h21] (h24) {$a^s_4$};
%	\node[inner sep=1pt, fit=(h21)(h24), draw] (h1box) {};
%	\node[right=0.2cm of h1box] {straight};
%
%	
%	\node[above=0.9cm of h21] (h1) {$a^l_1$};
%	\node[below right=0.12cm of h1] (h4) {$a^l_4$};
%	\node[inner sep=1pt, fit=(h1)(h4), draw] (hbox) {};
%	\node[right=0.2cm of hbox] {left};
%	
%
%	\node[below=0.9cm of h21] (h31) {$a^r_1$};
%	\node[below right=0.12cm of h31] (h34) {$a^r_4$};
	\node[below=0.3cm of feat2] (h3text) {\emph{objective movement values}};
%	\node[inner sep=1pt, fit=(h31)(h34), draw] (h3box) {};
%	\node[right=0.2cm of h3box] {right};

	\node[circle, draw, inner sep=2pt, right=1.5cm of gbox.east] (allact) {}; 

	\draw[->] (allact.east) edge[bend right=9] node[above, xshift=-2em] (lookupX) {lookup (3x)} (feat1.240);
	\draw[->] (allact.east) edge[bend right=9] node[above] (lookup) {} (feat1.305);
	\draw[->] (allact.east) edge[bend right=13] node[right] (lookup) {} (feat1.335);
	\draw[->] (gbox) edge node[above,align=center] {$\forall~f$\\$f$ closest} (allact);
	
	\draw[->, gray, dashed] (allact.east) edge[bend left=9] node[above, xshift=-2em] (lookup) {} (feat2.120);
	\draw[->, gray, dashed] (allact.east) edge[bend left=9] node[above] (lookup) {} (feat2.55);
	\draw[->, gray, dashed] (allact.east) edge[bend left=13] node[right] (lookup) {} (feat2.25);
	
	\node[above=0.18cm of lookupX] (qtables) {\color{red}Q-tables};

	\node[left=0.2cm of fbox, align=center] (obj) {\bf{$\forall$ objectives}\\e.g. \AVOID};

	\draw[->] (fbox) edge node[above] (select) {select closest} (gbox); 
	\node[above=0.6cm of select] (B) {human pos.};
 	
	\draw[->] (B) -- (select);
	\draw[<-] (lookupX) -- (qtables);
	
\end{tikzpicture}}
\label{Fig:ObtainMovementValues}}
\subfigure[The computation of distribution]{
\scalebox{0.8}{
	\begin{tikzpicture}
	
	\node[draw] (feat1) {$f_5: [q^{(5)}_l, q^{\scriptsize{(5)}}_s, q^{(5)}_r]$};
	
	\node[below=0.15cm of feat1, draw] (feat2) {$f_6: [q^{(6)}_l, q^{(6)}_s, q^{(6)}_r]$};
	\node[fit=(feat1)(feat2), draw] (boxavoid) {};
	
	\node[left=0.2cm of boxavoid] (h1text) {\AVOID};

	\node[below=0.4cm of feat2, draw] (feat3) {$f_2: [q^{(2)}_l, q^{\scriptsize{(2)}}_s, q^{(2)}_r]$};

	\node[fit=(feat3), draw] (boxfollow) {};
	
	\node[left=0.2cm of boxfollow] (h2text) {\FOLLOW};

	\node[below=0.4cm of feat3, draw] (feat4) {$f_8: [q^{(8)}_l, q^{\scriptsize{(8)}}_s, q^{(8)}_r]$};

	\node[fit=(feat4), draw] (boxcollect) {};
	
	\node[left=0.2cm of boxcollect] (h3text) {\COLLECT};

	\node[inner sep=0pt, right=0.4cm of boxfollow] (out) {};
	
	\draw[->] (boxavoid.0) -- (out);
	\draw[->] (boxfollow.0) -- (out);
	\draw[->] (boxcollect.0) -- (out);
	
	\node[right=2.5cm of out]  (vs) {$w_\AVOID \cdot [q^{(6)}_l, q^{(6)}_s, q^{(6)}_r]$};
	\node[above=0.02cm of vs,yshift=-0.2cm] {$+$}; 
	\node[below=0.02cm of vs,yshift=0.2cm] {$+$};
	\node[above=0.13cm of vs] (vl) {$w_\FOLLOW \cdot [q^{(2)}_l, q^{\scriptsize{(2)}}_s, q^{(2)}_r]$};
	\node[below=0.13cm of vs] (vr) {$w_\COLLECT \cdot [q^{(8)}_l, q^{\scriptsize{(8)}}_s, q^{(8)}_r]$};

	\node[above=0.2cm of vl.165] (weights) {\color{red}Weights};

	\node[right=of vs, align=center] (distr1) {probability\\distribution};
	\draw[->] (vs) -- node[above] {softmax} (distr1);

	\draw[->] (out) edge node[above] {$\forall$ combinations} node[below] {e.g. $f_6, f_2, f_8$} (vs);
	\draw[->] (weights) -- ([yshift=-0.15cm]vl.165);

	\end{tikzpicture}}\label{Fig:ObtainDistributions}
	}
	\caption{Sketched procedure to construct sets of distributions.}	
\label{Fig:DistributionProcedure}
\end{figure}

\paragraph{Movement-valuation.} Let us consider how to obtain movement values when combining objectives. These values are a kind of importance factor of the movement in a particular situation. The steps are outlined in Fig.~\ref{Fig:DistributionProcedure}. Fig.~\ref{Fig:ObtainMovementValues} outlines how to obtain values for the possible movements for each objective. These values are not always unique. Fig.~\ref{Fig:ObtainDistributions} describes how to combine the movement values for different objectives, and to translate it into distributions.
Details of the various steps are given below.
\paragraph{Relevant features.}
Recall that we consider three objectives, i.e.\ $\Objectives = \{\textsf{FOLLOW}, \textsf{AVOID},\textsf{COLLECT}\}$.
For each objective $o$,
we have exactly one \emph{corresponding feature-type} $f(o)$; Waypoints, obstacles and litter correspond to \textsf{FOLLOW}, \textsf{AVOID} and \textsf{COLLECT}, respectively.
As the behaviour of the human is independent of disappeared features, we call the set of present features the \emph{relevant features $\RelFeat_o(s) = \{ (f(o), \ell) \in \ActiveFeat \}$  w.r.t.\ objective $o$ in situation} $s= ((\humanloc, \humanangle), \ActiveFeat)$.%The set of all relevant features is given by $ \RelFeat(s) = \bigcup_{o \in \Objectives} \RelFeat_o(s)$.
\paragraph{Closest relevant features.}
We adopt the assumption from \cite{rothkopf2013modular} that for each objective $o$, only the closest feature of type $f(o)$ is relevant for the behaviour with respect to $o$.
\begin{remark}
While this assumption is strong, it reduces the number of hidden parameters a learning method has to estimate. Our method can be easily adapted to models where other/multiple features per type are relevant.
\end{remark}
Let $\close(s, o)$ be \emph{closest relevant features} for objective $o$ in situation $s$:
\[ \close(s, o) = \{f \in \RelFeat_o(s) \mid \forall f' \in \RelFeat_o(s).~ \dth(f) \leq \dth(f')  \}.\]
\begin{example}
	For the situation $s$ depicted in Fig.~\ref{Fig:ToyScenario}, 	$\close(s, \AVOID) = \{ f_5, f_6 \}$ and $\close(s, \FOLLOW) = \{ f_1 \}$.
\end{example}

While in the physical reality two objects are almost never equally far away, in the grid abstraction, this happens frequently. Thus, the set of closest relevant features is not necessarily a singleton. The human behaviour is underspecified in this case, and any of the features in the set might be the actually relevant feature.

While we support under-specification, it can be valuable to collapse this, and create the \emph{unique} closest relevant features. We do so by selecting the feature from the closest relevant features with (i) the smallest absolute angle, and in case of a tie (ii) left of the human over right-of-the human. This assumption allows us to treat larger benchmarks at the cost of a less precise model.
\begin{example}
Breaking the tie means that in the situation $s$ depicted in Fig.~\ref{Fig:ToyScenario}, \close($s$, \AVOID) = \{ $f_5$ \}, as $|\ath(f_5)| \approx \ang{26} < |\ath(f_6)| \approx \ang{63}$.
\end{example}

\begin{table}[t]

\caption{Illustrative fragment of Q-tables $Q_o^m(\gamma, d)$ with the bins for angle $\gamma_h$ (in degrees) and distance $\dth$ (in tiles) listed vertically and horizontally, respectively. Boldface entries correspond to the lowest (avoiding the movement the most) value over the different actions, while the coloured entries are used in Ex.~\ref{Ex:MovementValues}.}
\centering
\subtable[\AVOID, \HALEFT]{
\scalebox{0.76}{
\begin{tabular}{@{}lllll@{}}
\toprule
                               &  $[0,1]$                               &   $(1,2]$                              &    $(2,3]$                                &      $\hdots$                              \\ \midrule
\multicolumn{1}{l|}{$\hdots$}          & {\color[HTML]{C0C0C0} $\hdots$} & {\color[HTML]{C0C0C0} $\hdots$} & {\color[HTML]{C0C0C0} $\hdots$} & {\color[HTML]{C0C0C0} $\hdots$} \\
\multicolumn{1}{c|}{\textbf{}$[{-}90,{-}45)$} & \textbf{-0.5}                            & \textbf{-0.495}                   & \textbf{\color{green}-0.01\color{black}}                       & {\color[HTML]{C0C0C0} $\hdots$} \\
\multicolumn{1}{c|}{\textbf{}$[{-}45,{-}15)$} & \textbf{-1.21}                        & \textbf{-1.19}                    & \textbf{-0.59}                    & {\color[HTML]{C0C0C0} $\hdots$} \\
\multicolumn{1}{c|}{\textbf{}$[{-}15,{+}15]$} & -0.87                          & -1.12                    & -0.43                     & {\color[HTML]{C0C0C0} $\hdots$} \\
\multicolumn{1}{c|}{\textbf{}$({+}15, {+}45]$} & -0.75                           & -0.18                    & \color{orange}0\color{black}                                & {\color[HTML]{C0C0C0} $\hdots$} \\
\multicolumn{1}{c|}{\textbf{}$({+}45, {+}90]$} & \textbf{-0.75}                           & -0.06                    & 0                               & {\color[HTML]{C0C0C0} $\hdots$} \\
\multicolumn{1}{l|}{$\hdots$}          & {\color[HTML]{C0C0C0} $\hdots$} & {\color[HTML]{C0C0C0} $\hdots$} & {\color[HTML]{C0C0C0} $\hdots$} & {\color[HTML]{C0C0C0} $\hdots$} \\ \bottomrule
\end{tabular}
}
}
\subtable[\AVOID, \HASTRAIGHT]{
\scalebox{0.76}{
\begin{tabular}{@{}lllll@{}}
\toprule
                               &  $[0,1]$                               &   $(1,2]$                              &    $(2,3]$                                &      $\hdots$                              \\ \midrule
\multicolumn{1}{l|}{$\hdots$}          & {\color[HTML]{C0C0C0} $\hdots$} & {\color[HTML]{C0C0C0} $\hdots$} & {\color[HTML]{C0C0C0} $\hdots$} & {\color[HTML]{C0C0C0} $\hdots$} \\
\multicolumn{1}{c|}{\textbf{}$[{-}90,{-}45)$} & 0                            & 0                    & \color{green}0\color{black}                       & {\color[HTML]{C0C0C0} $\hdots$} \\
\multicolumn{1}{c|}{\textbf{}$[{-}45,{-}15)$} & -0.02                         & -0.96                    & -0.28                    & {\color[HTML]{C0C0C0} $\hdots$} \\
\multicolumn{1}{c|}{\textbf{}$[{-}15,{+}15]$} & -0.4                          & -1                    & \textbf{-1.12}                     & {\color[HTML]{C0C0C0} $\hdots$} \\
\multicolumn{1}{c|}{\textbf{}$({+}15, {+}45]$} & -0.27                           & -0.99                    & \color{orange}-0.67\color{black}                              & {\color[HTML]{C0C0C0} $\hdots$} \\
\multicolumn{1}{c|}{\textbf{}$({+}45, {+}90]$} & 0                           & 0                    & 0                               & {\color[HTML]{C0C0C0} $\hdots$} \\
\multicolumn{1}{l|}{$\hdots$}          & {\color[HTML]{C0C0C0} $\hdots$} & {\color[HTML]{C0C0C0} $\hdots$} & {\color[HTML]{C0C0C0} $\hdots$} & {\color[HTML]{C0C0C0} $\hdots$} \\ \bottomrule
\end{tabular}
}
}
\subtable[\AVOID, \HARIGHT]{
\scalebox{0.76}{
\begin{tabular}{@{}lllll@{}}
\toprule
                               &  $[0,1]$                               &   $(1,2]$                              &    $(2,3]$                                &      $\hdots$                              \\ \midrule
\multicolumn{1}{l|}{$\hdots$}          & {\color[HTML]{C0C0C0} $\hdots$} & {\color[HTML]{C0C0C0} $\hdots$} & {\color[HTML]{C0C0C0} $\hdots$} & {\color[HTML]{C0C0C0} $\hdots$} \\
\multicolumn{1}{c|}{\textbf{}$[{-}90,{-}45)$} & -0.96                            & -0                    & \color{green}0\color{black}                        & {\color[HTML]{C0C0C0} $\hdots$} \\
\multicolumn{1}{c|}{\textbf{}$[{-}45,{-}15)$} & -0.88                         & -0.77                    & 0                   & {\color[HTML]{C0C0C0} $\hdots$} \\
\multicolumn{1}{c|}{\textbf{}$[{-}15,{+}15]$} & \textbf{-0.88}                          & \textbf{-1.3}                    & -0.04                     & {\color[HTML]{C0C0C0} $\hdots$} \\
\multicolumn{1}{c|}{\textbf{}$({+}15, {+}45]$} & \textbf{-0.99}                          & \textbf{-0.99}                    & \textbf{\color{orange}-0.32\color{black}}                               & {\color[HTML]{C0C0C0} $\hdots$} \\
\multicolumn{1}{c|}{\textbf{}$({+}45, {+}90]$} & -0.5                           & \textbf{-0.13}                   & \textbf{-0.34}                               & {\color[HTML]{C0C0C0} $\hdots$} \\
\multicolumn{1}{l|}{$\hdots$}          & {\color[HTML]{C0C0C0} $\hdots$} & {\color[HTML]{C0C0C0} $\hdots$} & {\color[HTML]{C0C0C0} $\hdots$} & {\color[HTML]{C0C0C0} $\hdots$} \\ \bottomrule
\end{tabular}
}
}

\label{Tab:QTables}
\vspace{-0.5cm}
\end{table}

\paragraph{Movement-values.}
We assume that we have Q-tables as in~\cite{rothkopf2013modular}. 
In particular, for each \emph{movement} $m$ and each objective $o$, a Q-table $Q^m_o\colon \RR \times \RR_+ \rightarrow \RR$ maps the angle $\ath(f)$ and distance $\dth(f)$ between a human and a close relevant feature $f$ to an \emph{objective-movement-value}. 
Partial tables are given in Tab.~\ref{Tab:QTables}.
Alike to~\cite{rothkopf2013modular}, we translate the set of closest relevant features into a set of objective-movement-vectors $\objectivemovementValues{o}$ by a lookup in the Q-table, and store the feature for later use. 
\begin{align*}  
\objectivemovementValues{o} = & \{ (f, [q_l, q_s, q_r]) ~|~ f \in \close(s, o), & & q_l = Q_o^\HALEFT(\ath(f), \dth(f)), \\
&\quad  q_s =  Q_o^\HASTRAIGHT(\ath(f), \dth(f)),  & & q_r = Q_o^\HARIGHT(\ath(f), \dth(f))~\}
\end{align*}
The vector entries collect movement-values with respect to a fixed feature.
\begin{example}
\label{Ex:MovementValues}
For $s$ as depicted in Fig.~\ref{Fig:ToyScenario},
 $\close(s, \AVOID) = \{ f_5, f_6 \}$, we obtain by lookup in Tab.~\ref{Tab:QTables} -- using $\dth(f_i) = \sqrt{5} \in [2, 3)$ for $i \in \{5, 6\}$, $\ath(f_5) \approx \ang{26} \in [15, 45)$, and $\ath(f_6) \approx \ang{-63} \in [-90, -45)$ -- the following result:	
  $\objectivemovementValues{\AVOID} = \{ ( f_5 ,  [\color{orange}0, -0.67, -0.32\color{black}]), (f_6, [\color{green}-0.01, 0, 0\color{black}])	\}$.
\end{example}
\paragraph{Combining movement values.}
We assume that have an \emph{objective-weight vector} $w = [w_\AVOID, w_\COLLECT, w_\FOLLOW] \in \Distr(\Objectives)$ over the objectives, e.g.\ obtained by IRL as in~\cite{rothkopf2013modular}.
The objective-movement-vectors are translated into movement-vectors by calculating a weighted sum over all combinations: 
We first scale all $\objectivemovementValues{o}$ with $w_o$,
that is, the weighted-objective-movement values $\wobjectivemovementValues{o}$ = $\{ (f, w_o \cdot \vec{x}) | (f,\vec{x}) \in \objectivemovementValues{o} \}$. 
Then, we take the sum of the weighted vectors and construct the union of all involved features.
Formally, for any movement-vector and position of the human the corresponding set of movement-values, a pair consisting of the \emph{movement-vector} and a set of the features involved:
\[\movementValues = \{ (\{ f, f', f''\}, \vec{x} + \vec{x}' + \vec{x}'') | (f, \vec{x}, f', \vec{x}', f'', \vec{x}'')  \in \bigtimes_{o \in \Objectives} \wobjectivemovementValues{o} \}\]  where the sum is the component-wise sum.

\begin{example}
Similar to Ex.~\ref{Ex:MovementValues}, we obtain $\objectivemovementValues{\COLLECT} = \{(f_8, [2.5, 2.4, 2]) 	\}$,  $\objectivemovementValues{\FOLLOW} = \{(f_1, [0, 1.14, 0]) \}$.
The $\movementValues$ can be computed using the objective-weights (provided by e.g.\ IRL) $\vec{w} =   [0.414, 0.215, 0.369]$.
We get \begin{align*}
\movementValues = \{ &(\{ f_6, f_2, f_9 \}, 0.414 \cdot [.., .., ..] + 0.215 \cdot [.., .., ..] + 0.369 \cdot [.., .., ..]), \\
 	& (\{ f_7, f_2, f_9 \}, 0.414 \cdot [.., .., ..] + 0.215 \cdot [.., .., ..] + 0.369 \cdot [.., .., ..])\} \\
= \{ & (\{f_6, f_2, f_9\}, [0.53,  0.66,  0.3]), (\{f_7, f_2, f_9\}, [0.53, 0.93, 0.42] \} 
 \end{align*}
\end{example}

Notice that at the grid borders, only some movements are possible. Thus, we need to rule out such movements. In the course of this paper, we do this by resetting those movement-values to $-\infty$. If no movement is possible, we remove the vector from the movement values.
\paragraph{Distribution.}
The last step (cf.\ Fig.~\ref{Fig:ObtainDistributions}) is to transfer movement values into a distribution. Under the working hypothesis that the valuation reflects the likelihood that a human makes a specific move, we can translate this into a distribution over the movements -- provided any movement is possible. 

The stochastic behaviour of the human is obtained by translating any vector $\vec{x} \neq \vec{-\infty}$ for 
$(F, \vec{x}) \in \movementValues$ for some situation $s$ to a distribution over movements,
 by means of a \emph{softmax}-function~\cite{sutton1998reinforcement}: $\RR^n_\infty \rightarrow [0,1]^n$ -- which attributes most, but not all probability to the maximum, hence the name. Using  $e^{-\infty} := 0$, the distribution is defined as:
\[ \softmax_\temp(\vec{x})_i = \frac{e^{\nicefrac{\vec{x}_i}{\temp}}}{\sum_{i \leq |\vec{x}|} e^{\nicefrac{\vec{x}_i}{\temp}}}~. \]
For any invalid movement, the denominator is thus unaffected as the term is zero, the numerator is zero, effectively ruling out the transition.
The parameter $\temp$ is called the \emph{temperature}. Towards a zero temperature, the function is as a (hard) maximum, while with a high temperature it yields an almost uniform distribution. 
\subsection{MDP model of the human behaviour}
\label{Sec:FormalMDP}
Given the formal description above, we are now ready to construct the MDP for the human behaviour.
The state space is given by the set of situations. The initial state is given by the start-location of the human and the assumption that initially all features are present. 
%We take an arbitrary finite set of actions $A = \{ a_1, \hdots a_n \}$. 
%
As stated before we have two possible sources of non-determinism: 
(1)~Under-specification of the model: the closest relevant feature is not unique.
(2)~Insufficient confidence in some entries in the Q-table, e.g.\ if the amount of data does not allow us to draw conclusions.	
Here we only consider the first source, the latter is an extension with some more non-determinism.
 
\paragraph{Non-determinism due to under-specification.}
Resolving the non-singleton movement-vectors is modelled to be an action of the environment.
\begin{definition}
	The MDP $\MdpInit$ reflecting the human behaviour starting in $\humaninit$ on a environment $\Env = (\loc, \Feat)$ using temperature $\temp$ is given by 
	\begin{itemize}
		\item $\MdpStates = \{ (\humanpos, P) \mid \humanpos = (\humanloc, \humanangle) \in \loc \times \orientations, P \subseteq \Feat \}$
		\item $\sinit = (\humaninit, \Feat)$
		\item $\Act = \Feat^3$
		\item $\begin{aligned}[t]\probmdp(s, a) = \begin{cases} \{ \eff{{{(\humanActs)}_i}}(s) \mapsto \softmax_\temp(\vec{x})_i \mid i \in \{1,2,3\}  \} & (a,\vec{x}) \in \movementValues  \\
 			0 & \text{otherwise.}\\\end{cases}	
 \end{aligned}$
	\end{itemize}
\end{definition}
%\jpk{Make consistent with def1}

\paragraph{Rewards for the human performance.}
In order to evaluate the performance with respect to the objectives, we define \emph{transition-reward} mappings $\reward{o}$ for each $o \in \Objectives$, 
\begin{align*}
\forall s, s' \in \MdpStates~\forall a \in \Act, ~\reward{o}(s, a, s') =
\begin{cases}
 1  & \text{if }\RelFeat_o(s) \neq \RelFeat_o(s')\\
 0  & \text{otherwise.} 	
\end{cases}
\end{align*}
To define a combined reward, we want to avoid obstacles and collect litter, which means to penalise visiting some fields while rewarding others; this calls for the use of both positive and negative rewards, which is not possible in \prism. 
With the help of the goal-states, we can give rewards upon entering them and considering the performance afterwards\footnote{
In combination with the robot, this prevents distinguishing robot and human performance: here encoding the reward in the state space would be our last resort.}.

\subsection{SG model of human-robot interaction}
In order to obtain a model which models human behaviour and allows us to synthesise a plan for the robot, we have to consider a unified model. As we want to choose actions for the robot, the robot is naturally modelled as a (potentially non-probabilistic) MDP. Notice that the non-determinism of the robot is \emph{controllable}, whereas the non-determinism of the human model is \emph{uncontrollable}. This naturally leads to a stochastic two-player game. As a design choice, we let robot and human move in turns (typically alternating). While this abstraction is not inherently different from synchronous movements, this means that we can use single actions to determine the movements.

\paragraph{The robot model.} We can support any MDP on the same grid given as a \tool{PRISM}-module. Within the scope of the paper, we considered a simple robot either turning 90-degrees (left or right) in place, or moving forward. Synchronisation with the environment is via shared variables.

\paragraph{Considering the robot movement.} %Earlier, we already mentioned that any synchronization is done via a shared-memory -- i.e.\ shared variables -- approach. 
%Notice that in the combination with a robot movement, the grid is extended by a robot.
We both (1) considered a scenario where the human behaviour is not influenced by the robot and (2) the robot as an obstacle-feature. 
(1) yields a controller for the robot which is not intrusive, whereas (2) means that the human tries to avoid the robot and the controller takes this into account. For the latter, we had no suitable data and we assumed that the human treats the robot as if the robot were static.  However, the methodology presented here can be used as is if suitable Q-tables for dynamic obstacles are available.

\section{Experiments}
We developed a prototypical tool-chain to show the feasibility of our approach outlined in Fig.~\ref{fig:tool}.
The tool realizes Fig.~\ref{Fig:DistributionProcedure} for any scenario and exploits the \tool{PRISM-Games} v2.0beta3~\cite{DBLP:conf/tacas/KwiatkowskaPW16}.
We first discuss how to create the model and then present some evaluation results obtained under Debian 8 on a HP BL685C G7, 48 cores, 2.0GHz each, and 192GB of RAM.

%% \subsubsection{Implementation.}
\paragraph{State and transition encoding.}
We encode the SG for the joint human and (optional) robot behaviour in the \prism-language. 
The model has a module for the human and one for the robot. 
A global flag indicates whether the human (or robot) moves next. 
The robot (human) has a precondition that the robot (human) may move and an update to let the human (robot) move next.
% Schemes with multiple robot-moves per human-movement can be supported by means of a counter.
%% We include a given module for the robot (assuming some variable names). 
A module for the human consists of 3 integers to represent her location and orientation, and a boolean $b_f$ for each feature $f$, with $b_f$ true iff $f$ is present. 
The location of the (static) features are constants.
%Here, we discuss the scenario where the non-determinism is only due to underspecification.
%% JPK: this is obvious.  
%% While the state space of the SG is the product of the two modules, the size of the \prism-encoding equals the sum of the modules.
Though the set of reachable scenarios is exponential in the number of features, the encoding is cubic as the behaviour is based on the nearest present features only; see also App.~\ref{App:SizeOfEncoding}.

\paragraph{Reward encoding.}
The \prism language does not support state-action-target rewards as used in Sect.~3.3. 
We support two options for encoding the objective-reward:
%\begin{itemize}
(1)~Rescale the rewards to state-action rewards (preserving for expected reward measures~\cite{Put94}). As the rescaling depends on the probabilities, we de-facto have to scale rewards for each command which may reach a feature separately. Commands can only be named in the scenario without a robot; the combination of named commands and global variables (the status of the features) is not supported.
(2)~Introduce a first-time flag for any location with features indicating whether it is visited for the first time. This increases the number of states by $|\{ \ell ~|~ (\tp, \ell) \in \Feat \}| \cdot |\orientations|$. The number of commands does not increase, the BDD size is hardly affected. The rewards are attached to the states: the encoding is as large as the number of locations with features.
%\end{itemize}

\paragraph{Optimisations.}
We investigated various performance improvements. 
Two notable effective insights are:
(1) (Only) the Q-table for obstacle avoidance shows equal values for the far-away bins---indicating that human behaviour does not consider far-away obstacles. It is thus not necessary to distinguish which obstacle is nearest once they all induce the same lookup-value. This is especially relevant on large and sparse scenarios.
%  and in those situations where the robot location should be taken into account. 
(2) As every human location occurs on a large number of commands, naming expressions (called formulas in the \prism-language) reduces the overhead of specifying the location repeatedly. Using short non-descriptive variable names reduces the encoding further. Together, they reduce the parsing time by $40 \%$. 

%% \subsubsection{Evaluation.}
\paragraph{Model construction and parsing.}
As most case studies can be succinctly described in the \tool{PRISM}-language, parsing usually is not an issue.
Here it is.
Our models are up to 100,000 lines of \tool{PRISM} code.
App.~\ref{App:ModulBuildingPerformance} lists details about the model sizes and their building times. 
Parsing takes a significant amount of time. 
The hybrid \tool{PRISM} engine yields good overall performance: model construction is much faster than for explicit state spaces, and value iteration with many small probabilities ($< 0.01$) takes many iterations to converge, which is slow on a purely BDD-based engine. 

\paragraph{Evaluating the human.}
For the MDP model w/o robot, we compute some minimum and maximum probabilities; their difference indicates the relevance of the underspecification. 
Fig.~\ref{Fig:ExpEnvironment} gives some verification results for a 20x20 grid with 2 landmarks, 2 obstacles, and 7 waypoints. 
The module contains 84,000 commands.
Fig.~\ref{Fig:ExpVariances} plots the min/max probability to reach the goal area when the human is only told to follow the waypoints against the temperature (controlling the variability in the softmax function).
It shows that with low variability most humans indeed reach the other side without leaving the grid.
This quickly drops with higher variability (where any features are mostly ignored).
Fig.~\ref{Fig:BoundedSteps} indicates a similar behaviour for step-bounded reachability for different number of steps (x-axis) and temperatures 0.05 and 0.5.
Most humans need $>$ 30 steps to reach the goal, indicating that based on the given data, they very unlikely walk in straight lines. 
The gap due to underspecification is significant as long as the variability is not too high. 
With low variability, most humans arrive within 60 steps. 
Detailed analyses considering the obtained schedulers show where underspecification has the largest effect.
\pgfplotsset{every axis/.append style={
                    legend style={font=\tiny, at={((0.5,-0.3))}, align=left, anchor=north,draw=none ,mark size=2pt},
                    }}%
\begin{figure}[t]

\subfigure[Environment]{
\scalebox{0.14}{
\begin{tikzpicture}
\draw[step=1cm,gray,very thin, opacity=0.6] (0,0) grid (20,20);
\node[obstacle, scale=1.8] at (7,12) (f5) {};
\node[obstacle, scale=1.8] at (12,8) (f6) {};
\node[landmark, scale=1.8] at (7,8) (f8) {};
\node[landmark, scale=1.8] at (12,15) (f9) {};

\node[waypoint, scale=2.3] at (10,3) (f2) {};
\node[waypoint, scale=2.3] at (10,6) (f3) {};
\node[waypoint, scale=2.3] at (10,9) (f4) {};
\node[waypoint, scale=2.3] at (10,12) (f4) {};
\node[waypoint, scale=2.3] at (10,15) (f4) {};
\node[waypoint, scale=2.3] at (10,18) (f4) {};
\node[waypoint, scale=2.3] at (10,20) (f4) {};
\node[human={90}{green!80}, scale=2] at (10,0) (h) {};

\end{tikzpicture}
}
\label{Fig:ExpEnvironment}
}\hspace*{-0.9em}
\centering
\pgfplotsset{footnotesize}
\subfigure[Different variances]{
 \begin{tikzpicture}
\begin{axis}[width=3.9cm, height=3.3cm, ylabel={Prob}, xlabel={Temp.},
axis x line=bottom,mark size=0.8pt,
 axis y line=left,
 ymin = 0,
 ymax = 1,
 xmode=log,
 xtick={0.01, 0.1, 1, 2},
 xticklabels = {0.01, 0.1, 1, 2},
 x label style={font=\scriptsize,at={(axis description cs:0.8,0.25)},anchor=north},
 y label style={font=\scriptsize, at={(axis description cs:0.25,.5)},anchor=south},
% cycle list name=mycolor, 
 legend columns=2,
 legend entries={ $P_\text{min}(\eventually T)$, $P_\text{max}(\eventually T)$, $P_\text{min}(\eventually^{\leq 30} T)$, $P_\text{max}(\eventually^{\leq 30} T$)},
 legend image post style={scale=0.5},
% legend style={font=\tiny, at={(0,-0.3)}, anchor=center},
% legend pos=outer north east,
% legend to name=named,
legend cell align=left,
 yticklabel style={font=\tiny},
 xticklabel style={font=\tiny}]
]
\addplot+[smooth]
 table[x=Temperature,y=Pmin]
 {data/temp_variance.dat};
 \addplot+[smooth]
 table[x=Temperature,y=Pmax]
 {data/temp_variance.dat};
 \addplot+[smooth]
 table[x=Temperature,y=Pmin30]
 {data/temp_variance.dat};
 \addplot+[smooth]
 table[x=Temperature,y=Pmax30]
 {data/temp_variance.dat};
\end{axis}
\end{tikzpicture}
\label{Fig:ExpVariances}
}\hspace*{-0.9em}
\subfigure[Bounded steps]{
\begin{tikzpicture}
\begin{axis}[width=3.9cm, height=3.3cm, ylabel={Prob}, xlabel={Steps},
axis x line=bottom,mark size=0.8pt,
 axis y line=left,
 ymin = 0,
 ymax = 1,
 xtick={20, 40, 60},
 x label style={at={(axis description cs:0.8,0.25)},anchor=north},
 y label style={at={(axis description cs:0.25,.5)},anchor=south},
% cycle list name=mycolor, 
 legend columns=2,
 legend entries={ $P_\text{min}(\eventually^{\leq \#}T)$/.05, $P_\text{max}(\eventually^{\leq \#}T)$/.05, $P_\text{min}(\eventually^{\leq \#}T)$/.5, $P_\text{max}(\eventually^{\leq \#}T)$/.5},
  legend image post style={scale=0.5},
% legend to name=named2,
legend cell align=left,
 yticklabel style={font=\tiny},
 xticklabel style={font=\tiny}]
]
\addplot+[smooth]
 table[x=k,y=Pmin/0.05]
 {data/stepvariance.dat};
 \addplot+[smooth]
 table[x=k,y=Pmax/0.05]
 {data/stepvariance.dat};
 \addplot+[smooth]
 table[x=k,y=Pmin/0.5]
 {data/stepvariance.dat};
 \addplot+[smooth]
 table[x=k,y=Pmax/0.5]
 {data/stepvariance.dat};
\end{axis}
\end{tikzpicture}
\label{Fig:BoundedSteps}
}
\vspace*{-0.5cm}
\caption{Human performance evaluation.}
%\label{Fig:MCS11nperformance}
\vspace*{-0.4cm}
\end{figure}
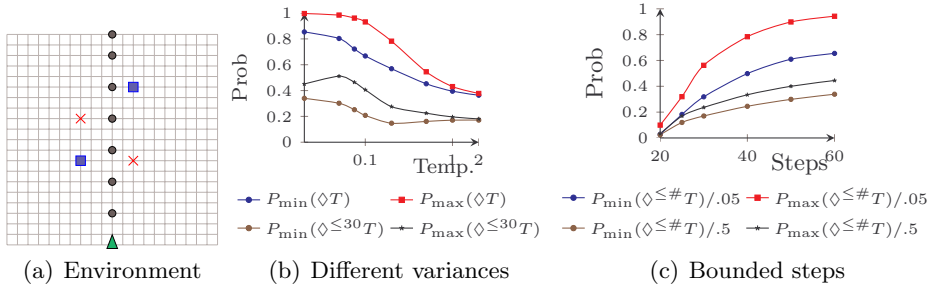

%% ---and counterexamples  uncovering the most relevant part of the reachable state space.
%To analyse any unexpected values, we can use two features:
%\begin{itemize}
%\item Analysis of the used scheduler: Modern model checkers allow us to obtain the minimising or maximising policy. By applying this to our model, we see what kind of choices are made in case of underspecification. This can help to refine the model in those places. 
%\item (High-level) counter examples: In order to get a smaller system than obtained by eliminating the non-determinism, the usage of counterexample computation helps to focus on a subset of the reachable state-space where the critical behaviour is rooted.
%\end{itemize}
 
\paragraph{Human-robot interaction.}
The state space is about a factor robot locations times robot directions (4) times turn-flag (2) larger. 
The number of choices is roughly twice the number of states. 
We built a $8{\times}8$ grid with 2/2/3 features; construction of the $1.6{\cdot}10^7$ states took 20 hours.
(Using the explicit engine, as \tool{PRISM-Games} has no symbolic engine.)
Model checking of a min/max reachability probability took 3 hours.
Constructing a $11{\times}11$ grid with 2/2/3 features with $7.4{\cdot}10^7$ SG-states took 198 hours. 
States are slowly created compared to typical benchmarks due to the large number of commands, rendering a thorough analysis impossible.
We made simplifications: (1) resolve the underspecification to have one player making trivial choices, or (2) consider a coalition where the underspecification is controllable. 
%% While this is suboptimal, it enables using a symbolic engine.
Both observations allow us to do a (suboptimal) analysis on the (isomorphic) single-player SG, i.e.\ an MDP. 
While this does not reduce the size of the model, it allows for using symbolic engines. 
We give figures for the coalition approach here; removing underspecification was slightly faster.
The aforementioned $11{\times}11$ grid-model takes $191,290$ MTBDD nodes ($13,725$ terminal) in the symbolic engine. 
Model construction in $19$ seconds.
Checking unbounded reachability takes about 500 iterations and four hours. 
As we are interested in the robot reaching its goal, finite-horizon plans are relevant. 
The first 100 iterations are done within 40 minutes and yield an almost identical scheduler.

\section{Conclusion and discussion}
We have successfully translated a cognitive model into a formal setting and used it to compute control plans for robots moving in the presence of humans handling complex tasks.
In the paper, we discussed the model as well as several (open) challenges we stumbled upon.

\paragraph{Lessons learned: the model.}
Based on the model-checking results, we obtained five lessons about the weighted Q-table model.
(1)~Fig.~\ref{Fig:BoundedSteps} shows that humans most likely walk in wavy lines. 
This is due to the lack of a \emph{notion of progress} in visiting waypoints---it does not penalise walking in circles, as only positive reward is earned  on visiting waypoints. 
Following a line and giving a penalty for any diverging move (as in~\cite{rothkopf2013modular}) would provide a notion of progress.
 %
% This leads to some artefacts in the model, as their is no to little incentive to turn towards the waypoints when they are further away. 
(2)~As the Q-tables do not \emph{take into account the border of the environment}, they are not avoiding a deadlock (or unspecified behaviour): The probability for leaving the grid is substantial in many cases.
(3)~For the discrete model, there is a potentially huge difference in behaviour based on how the underspecification is resolved; \emph{any analysis on the learned model has to take this underspecification into account}. 
(4)~Modelling variability over human behaviour by a single softmax and \emph{using a memoryless model} are rough estimates. Therefore it is quite likely that a human behaves inconsistent (in the SG) to statements in e.g.~\cite{matt_visiomotor}.
(5)~Finally, the Q-tables contain some unexpected \emph{outliers} which in some configurations lead to unexpected behaviour. 
%% Changing these values yields substantially different behaviour in several cases.  

\paragraph{Lessons learned: the method.}
The description for the human behaviour including variability can be translated into a formal model; allowing a variety of properties to be easily analysed---in particular, it allows for analysing under-specifications and ill-defined data. The generalisation to a robot planning scenario (and to a SG) is straightforward, and enables to compute plans fulfilling specific properties. Probabilistic verification of this model raised \emph{six challenges}:
(1)~The \emph{large number of different probabilities} in learned data blows up the encoding (and the BDD), it also prevents successful application of typical reduction techniques such as bisimulation.  It would be interesting to \emph{regularise the model} on the learning side, or use techniques like \emph{$\epsilon$-bisimulation} \cite{DBLP:conf/mfcs/BacciBLM13}. %%  to find more regular models (and estimate the error made).
(2)~The softmax function serves the only purpose of introducing variability; its nice features w.r.t. differentiability are not relevant here. \emph{Sensitivity analysis over $Q$-table values would be of interest} but current parameter synthesis techniques based on rational functions~\cite{param_sttt,dehnert-et-al-cav-2015} cannot cope with exponentials.
(3)~Despite the lacking of typical symmetries in the scenarios, its encoding is significantly smaller than enumerating all states, as (here) the behavior only depends on the nearest obstacles: even for three features of every kind, reduction of a factor over $50$ results.
(4)~While the approach yields promising results for the MDP scenario, the SG suffers from a state space explosion. Although the turn-based game is highly regular, this cannot be exploited by \emph{the lack of a symbolic engine} to solve SGs. 
(5)~The explicit engine suffers from the \emph{loss of information} in the \tool{PRISM}-encoding: formulas grouping states and making explicit that in a set of states only one or two commands are relevant are lost.
6)~Finally, the scheduler is always computed for the full state space, whereas based on reachability of many states, \emph{we are only interested in a fragment of the full state space}; for states with a low reachability probability, we can take any action.

\paragraph{Future work.}
For future work, we would also like to investigate automatic abstraction techniques as in \cite{WZH07} and restrict exploration of the model as in \cite{BCC+14}, as well as sensitivity analysis and/or model repair, potentially based on techniques presented in \cite{dehnert-et-al-cav-2015} or those in \cite{chen2013model}. 

{\small 
\par\smallskip\noindent\textbf{Acknowledgement.} We thank Mary Hayhoe and Matthew Tong for providing and explaining the data.
}

\bibliographystyle{splncs}
\bibliography{literature}
\appendix
\newpage

\section{The size of the encoding}
\label{App:SizeOfEncoding}
 We argue that the encoding is cubic.
For the description of the transition probabilities, it suffices to only consider under which circumstances a feature is not in $\close(s, o)$: That is exactly if there is another feature $f'$ of type $f(o)$ s.t. $\dth(f') < \dth(f)$. For any action $[f_\OBST, f_\LM, f_\WPT] \in \Feat_\OBST \times \Feat_\LM \times \Feat_\WPT$ and each human position, we have as a precondition $\bigland_{o \in \Objectives} \bigland_{f \in \Feat} \{ \neg b_f | f \in \Feat_o \land \dth(f_o) < f   \}$. That means that for each human position we have $\leq \prod_{o\in\Objectives} |\Feat_{f(o)}|$ commands.

If avoiding the robot is to be treated as a separate objective, than the transition encoding for any command in the human module has to be given separately for every possible location of the robot.

\newpage

\section{Experimental Performance}

\label{App:ModulBuildingPerformance}
Table~\ref{Tab:ModelBuildingHumanMDP} gives the grid size, the number of obstacles, landmarks, and waypoints, the number of commands, the number of states, choices and branches of the underlying model using $\temp=0.075$, the number of decision diagram nodes (with the number of terminal nodes in parentheses) as well as the time to parse and the time to build the model, either explicitly or via the symbolic engine. A $-$ indicates that the figure could not be obtained within 20 hours. The high number of terminal mtbdd nodes is a major challenge for the dd-based engines, whereas the explicit engine cannot cope with the large number of commands.

 In Table~\ref{Tab:ModelBuildingHumanSG} we give the same information for the models including a robot; notice that here, the symbolic numbers where obtained by transforming the model into an MDP. Please notice that the decision diagrams are comparatively small: This is due to the regularity of the alternating movements.

For the (8,8) SG models, the model checking time to obtain a scheduler maximising the probability to reach the other side of the grid without crashing, taking into account any scheduler resolving the underspecification is roughly 1300, 3000, 5400 and 10700 seconds, respectively.

\begin{table}[h]
\vspace{-.4cm}
\caption{Building times}
\centering
\subtable[Human MDP; non-determinism due to underspecification]{
\centering
\label{Tab:ModelBuildingHumanMDP}
\scalebox{0.9}{
\begin{tabular}{@{}rlrccccrrr@{}}
\toprule
grid 		& features & \# commands & states & choices & transitions & dd-nodes & $t_\text{parse}$ & $t^\text{exp}_\text{build}$ & $t_\text{build}^\text{sym}$ \\ \midrule
(11,11)     & 2/2/3    & 13325           &  9.4E4      & 2.4E5     & 9.9E4   &  131172 (13727)           &  7          &    965        &  11                     \\
(11,11)     & 4/4/4    &  33362          &  2.9E6      & 3.3E6     & 8.0E6    & 534607 (32540)             & 14           &     -       &  97                     \\
(16,16)     & 2/2/3    & 21256           &  1.1E5      & 1.2E5   & 3.1E5    &    135384 (16220)         &  27          &     2337       &    22                   \\
(16,16)     & 4/4/4    & 59527           &  3.5E6      & 4.0E6     & 1.0E7    & 414431 (34484)            & 41           &   -         & 194                      \\
(21,21)     & 2/2/3    & 42077           &  2.0E5      & 2.1E5     & 5.8E5  &    181376 (18236)         &  100         &   7192         &    41                  \\
(21,21)     & 4/4/4    & 95356           &  6.4E6      &  6.7E6    & 1.8E7  &  496709 (40441)           &  163          &    -        &  388                     \\ 
(21,21)     & 2/2/7    & 83900           &  3.2E6      & 3.4E6     & 9.3E6  &    466540 (30638)         &  135         &    -        &    310                  \\\bottomrule
\end{tabular}
}
}
\subtable[Human \& Robot 2p SG; non-determinism due to underspecification]{
\centering
\label{Tab:ModelBuildingHumanSG}
\scalebox{0.9}{
\begin{tabular}{@{}clrccccrrr@{}}
\toprule
grid 		& features & \# commands & states & choices & transitions & dd-nodes & $t_\text{parse}$ & $t_\text{build}^\text{exp}$ & $t_\text{build}^\text{sym}$ \\ \midrule
(8,8)     	& 1/1/2         & 2864           &  2.5E6      &    4.8E6      &  6.4E6           &  31097 (2511)  		&  3.0        	&  2181          &   2.5                    \\
(8,8)     &  1/1/3        & 3775           &    4.9E6    &    9.5E6     &      1.3E7        &  49580(3262)     		& 3.4    		&    5951        &  3.9                     \\
(8,8)     & 2/1/3         & 4464           &   8.4E6     &    1.6E7     &     2.2E7        &  72988 (4043)      & 3.5   &          14759  &   5.7                    \\
(8,8)     & 2/2/3         &  6624          &   1.6E7     &    3.2E7     &     4.4E7        & 128466 (6346) & 3.7   &          70595  &        10.1               \\
(11,11)     & 2/2/3         & 13179           & 7.4E7        &   1.4E8      &  2.0E9   & 191290 (13725)       &   7.5         &    7.1E5        &   19.2                    \\

(16,16)     & 2/2/7         & 50335           & 3.1E9        &   6.4E9      &  9.5E9   & 911575 (22870)       &      41.3      &     -       &   257                    \\
               \bottomrule
\end{tabular}
}
}
\end{table}
\noindent
These files are also available on \texttt{GitHub}: \url{https://github.com/moves-rwth/human_factor_models}.

\end{document}